\documentclass[11pt]{article} % For LaTeX2e
\usepackage{rldmsubmit,palatino}
\usepackage{graphicx}
\usepackage{comment} % enables the use of multi-line comments (\ifx \fi) 
\usepackage{fullpage} % changes the margin
\usepackage[noend]{algpseudocode}
\usepackage{algpseudocode}
\usepackage[colorinlistoftodos]{todonotes}
	\usepackage{amsmath}
\usepackage{amsfonts}
\usepackage{wrapfig}
\usepackage[export]{adjustbox}

\usepackage{algorithm}% http://ctan.org/pkg/algorithms
\usepackage{algpseudocode}% http://ctan.org/pkg/algorithmicx

\usepackage[papersize={8.5in,11in}]{geometry}

\title{Optimal Options for Multi-Task Reinforcement Learning Under Time Constraints}

\author{
Manuel Del Verme \\
Department of Computer, Automation, \\ and Management Engineering\\
University of Rome `Sapienza' \\
Rome, Italy \\
\texttt{manuel.delverme@gmail.com} \\
\And
Bruno Castro da Silva \\
Institute of Informatics\\
Federal University of Rio Grande do Sul (UFRGS)\\
Porto Alegre, Brazil\\
\texttt{bsilva@inf.ufrgs.br}\\
\And
Gianluca Baldassarre\\
Institute of Cognitive Sciences and Technologies  (ISTC)\\
National Research Council of Italy (CNR)\\
Rome , Italy\\
\texttt{gianluca.baldassarre@istc.cnr.it} \\
}

\begin{document}

\maketitle

\begin{abstract}
Reinforcement learning can greatly benefit from the use of options as a way of encoding recurring behaviours and to foster exploration.
An important open problem is how can an agent autonomously learn useful options when solving particular distributions of related tasks.
We investigate some of the conditions that influence optimality of options, in settings where agents have a limited time budget for learning each task and the task distribution might involve problems with different levels of similarity.
We directly search for optimal option sets and show that the discovered options significantly differ depending on factors such as the available learning time budget and that the found options outperform popular option-generation heuristics.
\end{abstract}
\acknowledgements{This project has received funding from the European
Union's Horizon 2020 Research and Innovation Program under
Grant Agreement no. 713010 (GOAL-Robots -- Goal-based
Open-ended Autonomous Learning Robots). This work was also partially supported by the Brazilian FAPERGS under grant no. 17/2551-000.}

\startmain % to start the main 1-4 pages of the submission.
\section{Introduction} 
Reinforcement learning (RL) is widely used to train autonomous agents with little human feedback \cite{SuttonBarto2018ReinforcementLearningAnIntroduction}.
However, even to learn to solve simple tasks it can require millions of interactions.
A promising approach to improve the learning speed relies on the \textit{options framework} \cite{Sutton1999}
An option is a `chunk of behaviour' that is formally defined as an \textit{initiation set}, establishing in which states the option is available;
a \textit{policy}, indicating which actions to perform in each state; and a \textit{termination condition}, establishing when the option execution is terminated.
RL systems can benefit from the use of options to support faster exploration and learning especially when rewards are sparse or when the solution to a problem involves recurring behaviours.

%The overall problem
An important open problem is how can an agent autonomously learn options that are useful to solve tasks drawn from a given task distribution. Recent approaches have searched options for specific optimisation problems but they have not studied how optimal options are affected by different task features such as limited learning time budgets, task rewards, initial states, and the learning algorithm used.
%
%\subsection{Extracting options from the transition function}
Various heuristics based on state transitions have been proposed to self-generate options, for example \textit{bottleneck} options and \textit{betweenness} \cite{SimsekBarto2008Skillcharacterizationbasedonbetweenness}, or \textit{eigenoptions} \cite{LiuMachadoTesauroCampbell2017TheEigenoptionCriticFramework}.
However, it is not always possible to determine \textit{a priori} which heuristic will be appropriate for a particular type of tasks and constraints.
Other approaches 
\cite{BaconHarbPrecup2017TheOptionCriticArchitecture,FransHoChenAbbeelSchulman2017MetaLearningSharedHierarchies}
% \cite{BaconHarbPrecup2017TheOptionCriticArchitecture,FransHoChenAbbeelSchulman2017MetaLearningSharedHierarchies, DBLP:journals/corr/abs-1802-06070}
interleave the problem of finding option policies and a policy over the found options. While these approaches can capture regularities across tasks they do not consider other important factors.

%Tasks with different time budgets
Here we study how optimal options might depend on a very important feature of the tasks to be solved, namely the time budget
%(i.e. the number of samples from the environment)
available for learning each task.
To this purpose, we used exhaustive searches over given spaces of option sets to find the optimal options for tasks sampled from a given distribution.
Although this approach does not scale to large problems, it allows us to show how optimised option sets vary as a function of the available learning time budget and that popular approaches to option creation often remain sub-optimal in some conditions.

%Richieste da un reviewer di ERL2018 (paper rigettato)
% \cite{Yao:2014:UOM:2968826.2968937,2017arXiv171011089M}

%\section{Problem formulation}
 \section{Methods}
\label{sec:methods}
 \subsection{Problem formulation}
% $\label{sec:problem}
We consider an agent that at each time $t$ can perform a selected primitive action from some action space $A$ when operating in an environment with state space $S$ and transition function $P(s'|s,a)$. 
We also consider a distribution $P(\tau)$ of \textit{tasks} $\tau$ that the agent has to solve. Tasks are Markov Decision Processes (MDP) sharing the same states, actions, and transition function, but they differ for having a different reward function $r_\tau(s,a)$.
We assume that the agent has a time budget $C$ to learn a policy $\pi_C(a|s,\tau)$  for each task $\tau$ (denoted as $\pi_C(\tau)$ with a simplified notation).

We consider agents that may use options alongside primitive actions and assume that each option $o$ can be selected at any state and that executing it corresponds to following its policy $\pi_o(a|s)$ for a random amount of steps until it reaches a termination state $g_o$ (`goal').
Our search algorithm exhaustively explores the whole option space $\mathcal{O}$ to identify the option set $O$, formed by a number $\omega$ of options, that maximises the expected performance $J(O,C)$ over the task distribution $P(\tau)$:
ption
\begin{eqnarray}
\max_{O \in \mathcal{P}(\mathcal{O})}
J(O,C)=
  \max_{O \in \mathcal{P}(\mathcal{O})} \int P(\tau) J(\pi_{O,C}(\tau), \tau) d \tau
  \label{eq:objective_with_options}
\end{eqnarray}
where $\mathcal{P}(\cdot)$ is the space of possible option sets; $\pi_{O,C}(\tau)$ is the policy acquired by a given RL algorithm under time budget $C$ for solving task $\tau$ by using both primitive actions $A$ and options $O$;
and $J(\pi_{O,C}(\tau), \tau)$ is the expected performance of $\pi_{O,C}(\tau)$ for task $\tau$ over a time horizon $M$:
\begin{eqnarray}
  && J(\pi_{O,C}(\tau), \tau) = E\left[\sum_{t=1}^M \left(r_t | \pi_{O,C}(\tau), \tau \right) \right]
  \label{eq:J_theorical}
\end{eqnarray}
where the expectation is over trajectories resulting from deploying policy $\pi_(O,C)$ in task $\tau$, the initial state distribution of $\tau$, and the stochasticity of the environment. 
The maximisation in Eq.~\ref{eq:objective_with_options} also depends on the RL algorithm that is assumed to be given and fixed.

%\subsection{Search and evaluation of option sets}
%\label{sec:strategies}

\begin{figure}
    \noindent\begin{minipage}[l]{0.35\textwidth}
    \centering
     \includegraphics[width=0.8\textwidth,frame]{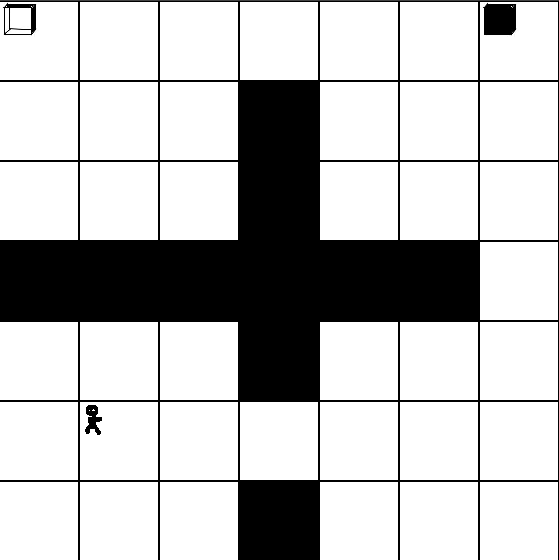}
        \caption{Food-source domain. The grid-world has walls (black), and food sources, here located top left (empty) and top-right (full).}
        \label{fig:food_domain}
    \end{minipage}
\noindent\begin{minipage}[r]{0.65\textwidth}
  \begin{algorithm}[H]
    \caption{Best Option Set Search (BOSS)}
    \label{alg:BOSS}
    \begin{algorithmic}
            \Function{BOSS}{$\{O_i\}_{i\in[1,\ldots, N]}$} \Comment{$\{O_i\}_{i\in[1,\ldots, N]}$ : option sets to be evaluated, each one with a number $\omega$ of options}
            \For{$k\in[1,\ldots, K]$}
               \State Sample task $\tau_k \sim P(\tau)$
               \For{each option set $\{O_i\}_{i\in[1,\ldots,N]}$}
                 \State Learn policy $\pi_{O_i,C}(\tau_k)$ based on  $O_i$ and $A$
                 \State Evaluate $\pi_{O_i,C}(\tau_k)$ performance $\hat{J}(\pi_{O_i,C}(\tau_k),\tau_k)$
             \EndFor
            \EndFor
            %\For{$i \in [1\ldots N]$}
            \State Compute  $\hat{J}(O_i,C)=\frac{1}{K} \sum_k \hat{J}(\pi_{O_i,C}(\tau_k),\tau_k) \ \forall i \in [1,...,N]$
            %\EndFor
            \State \textbf{return} $\arg\max_{i\in[1,\ldots, N]} \hat{J}(O_i,C)$
            \EndFunction
    \end{algorithmic}
  \end{algorithm}
\end{minipage}
\end{figure}

Since evaluating Eq.~\ref{eq:objective_with_options}
would be computationally unfeasible,
%would require to measure the performance of policies over an infinite number of tasks,
we instead estimate the quality of each option set as follows (Algorithm \ref{alg:BOSS}). First we sample a set of $K$ tasks $\{\tau_k\}_{k\in[1,\ldots,K]}$ from the distribution $P(\tau)$.
For each task $\tau_k$, the chosen RL algorithm (here, Q-learning \cite{WatkinsDayan1992Qlearning}) uses a given candidate option set $O_i$, together with the primitive actions $A$, to learn a policy $\pi_{O_i,C}(\tau_k)$ for task $\tau_k$ under learning time budget $C$.
The acquired policy is then empirically evaluated multiple times on task $\tau_k$ to obtain a low-variance estimate of its performance, denoted as $\hat{J}(\pi_{O_i,C}(\tau_k),\tau_k)$.
The overall performance of each option set $O_i$ is then computed as the average performance over  $K$ tasks drawn from the distribution: $\hat{J}(O_i,C)=\frac{1}{K} \sum_k \hat{J}(\pi_{O_i,C}(\tau_k),\tau_k)$. 

\subsection{Domain and option space}
\label{sec:setup}

%Environment/domain
We first consider tasks corresponding to different MDPs defined over a base grid-world domain with 39 cells (cf. \cite{SinghLewisBarto2009Wheredorewardscomefrom}; Fig~\ref{fig:food_domain}). 
Each task corresponds to a Markov Decision Problem (MDP) where two food sources are located in non-overlapping positions in two randomly chosen corners of the environment, thus generating (12 different possible tasks).
%
%Output
The agent has four  actions, $\{north, east, west, south\}$, and moves one cell in a given direction only if it selects the corresponding primitive action two times in a row; for example $\langle east, east \rangle$ moves the agent 1 cell to the right.
This feature, that is inspired by the fact that `movements' in animals and robots often correspond to sequences of primitive actions and captures in an abstract fashion the distinction between regularities involving the agent's body and the environment, is useful to magnify the effect of good options.
Even when a correct action sequence is performed (e.g. $\langle east, east\rangle $), it can still fail to move the agent in the intended direction with a $10\%$ probability.%, thus introducing stochasticity in the environment.

%Input
The state of the agent is described by a scalar indicating its cell position, the state of the two food sources (present/absent) and the last performed primitive action.
The number of the possible states is, therefore, 
$49\ positions\ \times\ 2\ source\_states\ \times\ 4\ possible\_actions =\ 392$.
%
%Reward function, food replacement
As the agent moves in the environment, it incurs in a reward of $-1$ at each time step, and $+100$ if it eat a food by stepping on it.
When a food is eaten the source is emptied and the other food source is refilled.

%%%%%%%%%%%%%%%%%%%%%%%%%%%%%%%%%%
%\subsection{Option Representation}
%\label{sec:option_set}

In our experiments we allowed two types of options:
%\begin{enumerate}
%\item 
(1) Two-step lower-level options corresponding to the sequence of two primitive actions recreating the canonical movement actions (north, south, east, west).
%\item 
(2) Variable length options that terminate in a particular goal state $g_t$ representing reachable goal cells in the environment.
%\end{enumerate}
In our experiments, we consider option sets composed of a number $\omega \in \{0,1,2\}$ options and all the low-level options.
We compare our searched options with a pre-defined high-level option set corresponding to bottleneck states, in particular, these 3 high-level options with goals located at the 3 `doorways' in the environment.

%\section{Performance Evaluation}
%\label{sec:performance}
The considered domain involves a distribution $P(\tau)$ of 12 tasks where one task $\tau \sim P(\tau)$ is defined by a certain food source configuration. To reduce variability in the evaluation rather than sampling we tested every option set over all the tasks $\{\tau_k\}_{k\in [1,\ldots,12]}$.
The policy of each option $O_i$ was searched with Q-learning \cite{WatkinsDayan1992Qlearning}.
For each task $\tau_k$, the $\omega$ options in $O_i$ were then used, together with the primitive actions $a \in A$, to learn with a RL algorithm (again Q-learning) the policy to solve the task. 
The initial state distribution was uniform and each episode terminated after 500 steps or after the agent ate 3 foods. 
This learning process was constrained by a learning time budget, lasting $C$ time steps, and this lead to learn the policy $\pi_{O_i,C}(\tau_k)$.
$\hat{J}(\pi_{O_i,C}(\tau_k),\tau_k) = \frac{1}{40} \sum_{s\in S_0} \left[\sum_{t=1}^D  r_t| \pi_{O_i,C}(\tau_k), \tau_k, s_0 = s\right]$. %
The performance of $O_i$ was then computed as:
$\hat{J}(O_i,C)=\frac{1}{12} \sum_{k=1}^{12} \hat{J}(\pi_{O_i,C}(\tau_k),\tau_k)$.

%%%%%%%%%%%%%%%%%%%%%%%%%%%%%%%%%%
% \begin{figure}[htb!]  
\begin{wrapfigure}[21]{r}{0.5\textwidth}
\label{fig:performance}

% \centering
    %(a)
    % \includegraphics[width=0.65\columnwidth]{performance_complex.png}
    % \vspace{-75pt}
    \includegraphics[width=0.5\textwidth]{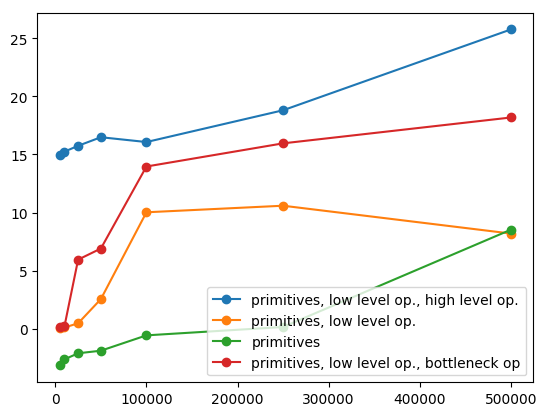}
    % (b)\includegraphics[width=0.8\columnwidth]{performance_mines.png}
    \caption{Performance (y axis) time budgets (x-axis) with: (1) low and high level options; (2) low-level options; (3) primitive actions only; (4) low-level and bottleneck options.}
    % \vspace{-75pt}
    \end{wrapfigure}
% \end{figure}The score of each option set was computed as follows. The performance of a given policy using the option-set for task $\tau_k$ was estimated by evaluating it in 40 tests, each one starting with the agent located in one of the 40 non-wall cells $S_0$. The performance of each test was defined as the number of foods collected in $D=1000$ time steps, and the total performance of the policy was the average over the tests:
%
\section{Results: learning optimal options under time constraints}
\label{sec:performance_BOS}

Fig.~2
%~\ref{fig:performance}
shows that, as expected, an agent with only primitive actions has worse performance.
Instead, even low complexity short-lasting option are beneficial with low time budgets.
Interestingly, popular option sets such as those based on \textit{bottlenecks} perform remarkably badly when used with low time budget, and improve only when the budget increases.

\begin{figure}%[htb!]
\centering
    (a)\includegraphics[width=0.3\columnwidth]{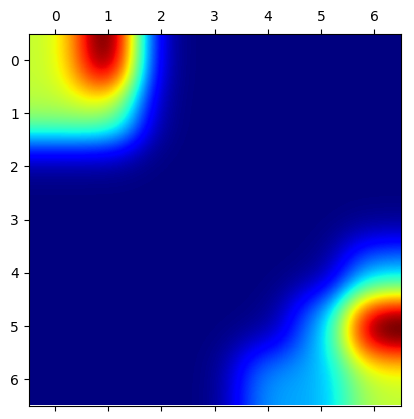}
    (b)\includegraphics[width=0.3\columnwidth]{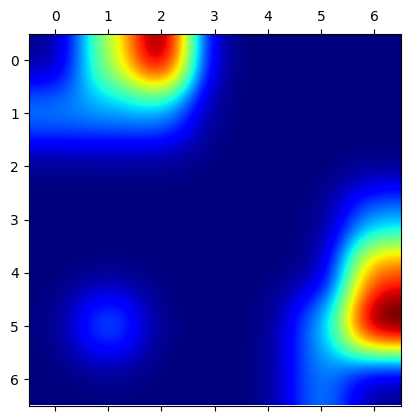}
    (c)\includegraphics[width=0.3\columnwidth]{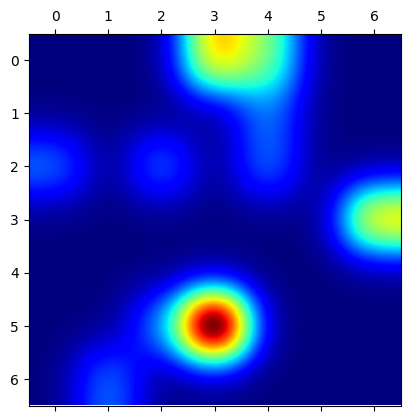}
    \caption{
Distribution of high-level goal locations for the top 10 option sets ($\omega=2$) given different learning time budgets: (a) 5,000 steps; (b) 25,000 steps; (c) 500,000 steps.}
    \label{fig:goal_distribution}
    \vspace{-20pt}
\end{figure}

Fig.~\ref{fig:goal_distribution} shows the distribution of the high-level navigation options of the top ten option sets found with Algorithm \ref{alg:BOSS}.
These best options depend on both the learning time budget available and some features of the environment.
Fig.~\ref{fig:goal_distribution}a shows that for a low time budget (C=5,000), the agent `bets' on some of the corners where food might be located and this strategy is successful for at least some of the 12 tasks with food sources located at some specific two corners.
Fig.~\ref{fig:goal_distribution}b shows that with intermediate time budgets, the found options are employed to move the agent towards locations that are close to the corners, but not on them: this facilitates the navigation between all rooms of the environment while also allowing a rapid reach of 2 of the 4 corners. Fig.~\ref{fig:goal_distribution}c shows that for higher time budgets, the best options correspond to bottlenecks options. 
Moreover, the system always chooses the bottleneck between the two south rooms. 
The reason is that navigating between those two rooms is particularly difficult since the agent has no walls around it to constrain possible movements.
\section{Conclusions}
\label{sec:conclusions}
We have empirically shown that there is no absolutely optimal set of options but the best set varies drastically with a given learning time budget.
This suggests that options generated by fixed heuristics based on the structure of the environment (e.g., bottleneck options) may be sub-optimal in some settings.
Other elements of tasks and settings, such as the reward functions, the initial states, and the learning algorithm used, might also affect the optimal options.

\bibliography{1_bibliography}
\bibliographystyle{abbrv}
\end{document}